\documentclass[runningheads]{llncs}

\usepackage{graphicx}
\usepackage{amsmath}
\usepackage{cite}  
\usepackage{graphicx}
\usepackage{makecell}
\usepackage{amsfonts}
\usepackage[caption=false]{subfig}
\usepackage{multirow}
\usepackage{amssymb}
\usepackage[misc,geometry]{ifsym}
\usepackage{booktabs}

\usepackage{color}

\usepackage{multirow}
\usepackage{url}
\newcommand{\etal}{\emph{et~al.}}

\begin{document}
\title{Cross-modal Attention for MRI and Ultrasound Volume Registration}
%
%

\author{Xinrui Song\inst{1}\and Hengtao Guo\inst{1}\and Xuanang Xu\inst{1}\and Hanqing Chao\inst{1}\and Sheng Xu\inst{2}\and Baris~Turkbey\inst{3}\and Bradford J. Wood\inst{2}\and Ge Wang\inst{1}\and Pingkun Yan\Letter\inst{1}}


\authorrunning{---}
\institute{Department of Biomedical Engineering and Center for Biotechnology and Interdisciplinary Studies, Rensselaer Polytechnic Institute, Troy, NY 12180, USA\\
\email{yanp2@rpi.edu}\\
\and
Center for Interventional Oncology, Radiology \& Imaging Sciences, National Institutes of Health, Bethesda, MD 20892, USA\\
\and
Molecular Imaging Program, National Cancer Institute, National Institutes of Health, Bethesda, MD 20892, USA}
%
\maketitle              

\begin{abstract}

Prostate cancer biopsy benefits from accurate fusion of transrectal ultrasound (TRUS) and magnetic resonance (MR) images. In the past few years, convolutional neural networks (CNNs) have been proved powerful in extracting image features crucial for image registration. However, challenging applications and recent advances in computer vision suggest that CNNs are quite limited in its ability to understand spatial correspondence between features, a task in which the self-attention mechanism excels. This paper aims to develop a self-attention mechanism specifically for cross-modal image registration. Our proposed cross-modal attention block effectively maps  each of the features in one volume to all features in the corresponding volume. Our experimental results demonstrate that a CNN network designed with the cross-modal attention block embedded outperforms an advanced CNN network 10 times of its size. We also incorporated visualization techniques to improve the interpretability of our network. The source code of our work is available at \url{https://github.com/DIAL-RPI/Attention-Reg}.

\end{abstract}

\section{Introduction}

Image-guided interventional procedures often require registering multi-modal images to visualize and analyze complementary information. For example, prostate cancer biopsy benefits from fusing transrectal ultrasound (TRUS) imaging with magnetic resonance imaging (MRI) to optimize targeted biopsy. However, image registration is a challenging task especially for multi-modal images.
Traditional multi-modal image registration relies on maximizing the mutual information between images \cite{maes1997multimodality,wells1996multi}, which performs poorly when the input images have complex textural patterns, such as in the case of MRI and ultrasound registration.
Feature based methods compute the similarity between images by representing image appearances using features \cite{HeinrichMIND}. However, feature engineering limits the registration performance on images in different contrasts, of complicated features, and/or with strong noise.

In the past several years, deep learning has become a powerful tool for medical image registration, starting from the early works of using neural networks for similarity metric computation to direct transformation estimation~\cite{wu2013unsupervised,haskins2020deep}. 
For example, Haskins \etal~\cite{haskins2019learning} developed a deep learning metric to measure the similarity between MRI and TRUS volumes. The correspondences between the volumes is established by optimizing the similarity iteratively, which can be computationally intensive.
de~Vos \etal~\cite{de2017end} proposed an end-to-end unsupervised image registration method to train a spatial transform network by maximizing the normalized cross correlation. Their method can directly estimate an image transformation for registration.
Balakrishnan \etal~\cite{balakrishnan2019voxelmorph} further used mean squared voxel-wise difference and local cross-correlation to train a registration network to map image features to a spatial transformation. 
While the way of estimating such an image transform underwent major changes, researchers also developed novel ways to supervise the network learning process. Hu \etal~\cite{hu2018weakly} trained an image registration framework in a weakly supervised fashion by minimizing the differences between segmentation labels of the fixed image and a warped moving image.
Yan \etal~\cite{yan2018adversarial} developed an adversarial registration framework using a discriminator to supervise the registration estimator.

The aforementioned deep learning methods map the composite features from input images directly into a spatial transformation to align them. So far, the success comes from two primary sources. One is the ability of automatically learning image representations through training a properly designed network. The other is the capability of mapping complex patterns to an image transformation. The current methods mix these two components together for image registration. However, converting image features to a spatial relationship is extremely challenging and highly data-dependent, which is the bottleneck for further improvements of the registration performance.

\begin{figure}[t]
\centering
	\includegraphics[width=.9\textwidth]{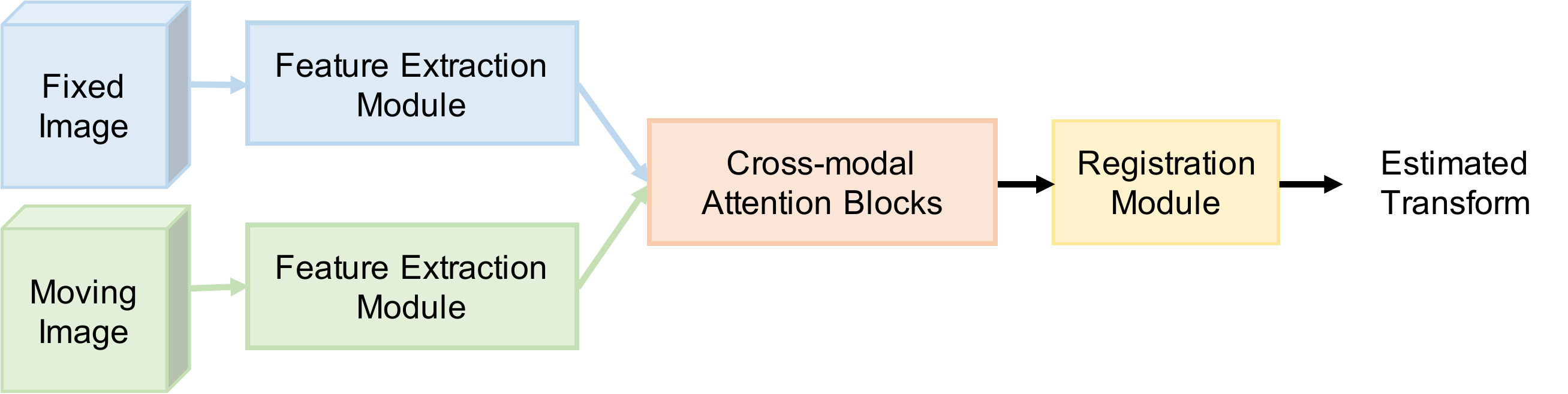}
	\caption{{Overview of the proposed registration framework with cross-modal attention. }} 
	\label{fig:overview}
\end{figure}

In this paper, we propose a novel cross-modal attention mechanism to explicitly use the spatial correspondence to improve the performance of neural networks for image registration. By extending the non-local attention mechanism~\cite{wang2018non} to an attention operation between two images, we designed a cross-modal attention block that is specifically oriented towards registration tasks. The attention block captures both local features and their global correspondence efficiently. Embedding this cross-modal attention block into an image registration network, as shown in Fig.~\ref{fig:overview}, improves deep learning based multi-modal image registration, attaining both feature learning and correspondence establishment explicitly and synergically.

By adding the cross-modal feature correspondence, the image registration network can achieve better registration performance with a much simpler architecture. To the best of our knowledge, this is the first work to embed the non-local attention in the deep neural network for image registration. 

In our experiments, we demonstrate the proposed method on the 3D MRI-TRUS fusion task, which is a very challenging cross-modality image registration problem. The proposed network was trained and tested on a dataset of 650 MRI and TRUS volume pairs. The results show that our network significantly reduced the registration error from $10.17 \pm 5.75$mm to $3.71 \pm 1.99$mm. The proposed method also outperformed state-of-the-art methods with only 1/10 to 1/5 of the number of parameters used by the competitors, as well as significantly reduced the run time.

\section{Method}

In this image registration application, the MRI volume is considered to be the fixed image, and the TRUS volume is the moving image. Our registration network consists of three main parts, as shown in Fig.~\ref{fig:overview}. The feature extractor uses convolutional and max pooling layers to capture regional features, and down samples the input volume. Then we use the proposed cross-modal attention block to capture both local features and their global correspondence between modalities. Finally, this information is fed to the deep registrator that further fuses information from two modalities and infers the registration parameters.

\begin{figure}[t]
\centering
	\includegraphics[width=.7\textwidth]{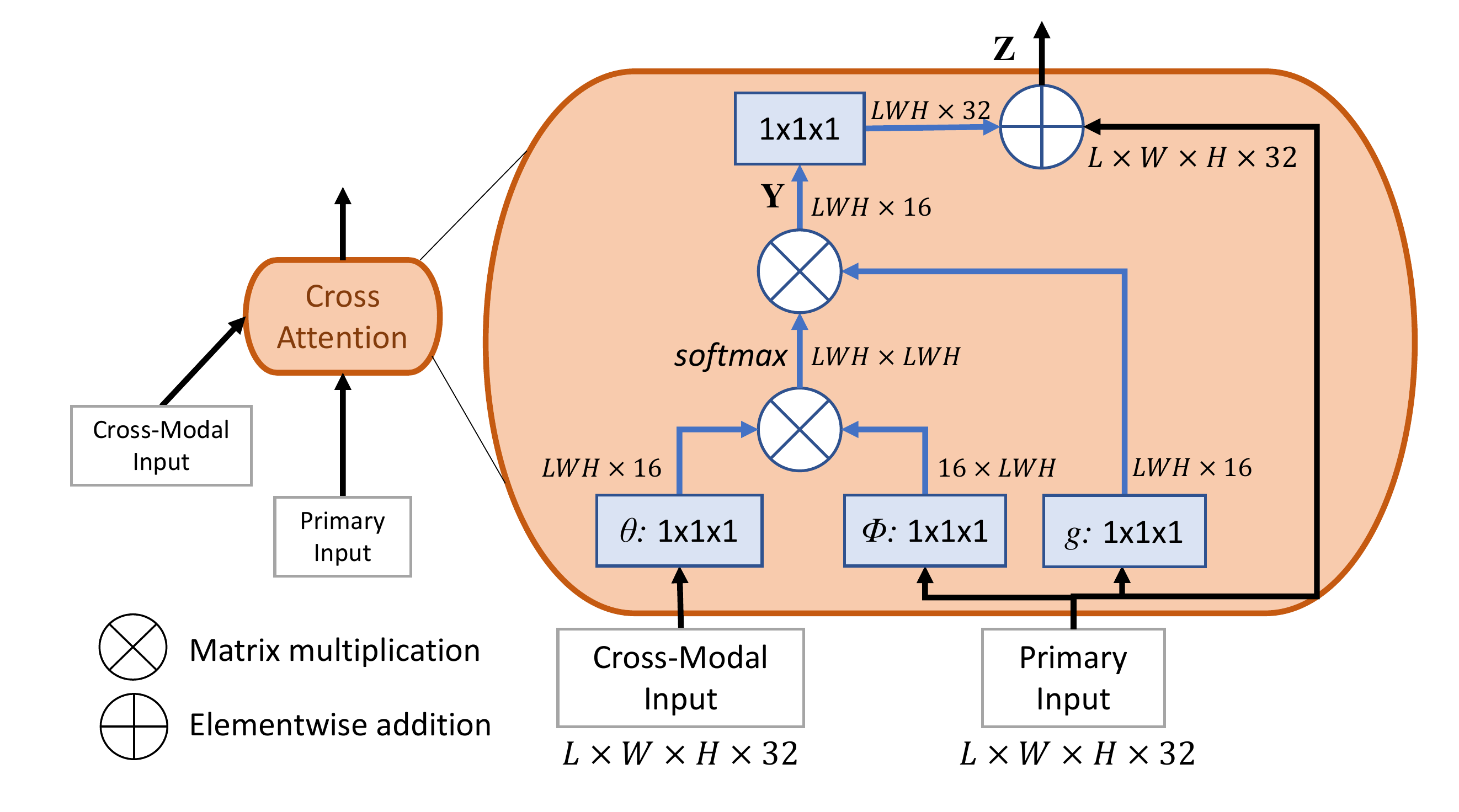}
	\caption{The proposed cross-modal attention block.}
	\label{fig:attention}
\end{figure}

\subsection{Cross-modal Attention}

The proposed cross-modal attention block takes image features extracted from MRI and TRUS volumes by the preceding convolutional layers. Unlike the non-local block~\cite{wang2018non} computing self-attention on a single image, the proposed cross-modal attention block aims to establish spatial correspondences between features from two images in different modalities. Fig.~\ref{fig:attention} shows the structure of the proposed cross-modal attention block.
The two input feature maps of the block are denoted as primary input $P \in \mathbb{R}^{LWH \times 32}$ and cross-modal input $C\in \mathbb{R}^{LWH \times 32}$, respectively. $LWH$ indicates the size of each 3D feature channel after flattening. The block computes the cross-modal feature attention as
\begin{equation}
\label{eq:attention}
\mathbf{y}_i = \frac{\sum_{\forall j}f(\theta(\mathbf{c}_i)^T\phi(\mathbf{p}_j))g(\mathbf{p}_j)}{\sum_{\forall j}f(\theta(\mathbf{c}_i)^T\phi(\mathbf{p}_j))},
\end{equation}
where $\mathbf{c}_i$ and $\mathbf{p}_j$ are features from $\mathbf{C}$ and $\mathbf{P}$ at location $i$ and $j$, $\theta(\cdot)$, $\phi(\cdot)$ and $g(\cdot)$ are all linear embeddings, and $f(\cdot)=\exp(\cdot)$. In Eq.~\ref{eq:attention}, $f(\cdot)$ computes a scalar representing correlations between the features of these two locations, $\mathbf{c}_i$ and $\mathbf{p}_j$. The result $\mathbf{y}_i$ is a normalized summary of features on all locations of $\mathbf{P}$ weighted by their correlations with the cross-modal feature on location $i$. Thus, the matrix $\mathbf{Y}$ composed by $\mathbf{y}_i$ integrated non-local information from $\mathbf{P}$ to every position in $\mathbf{C}$. Finally, the block's output $\mathbf{Z}$ is the sum of $\mathbf{Y}$ and $\mathbf{P}$ to allow efficient back-propagation. Therefore, the feature of a location $k$ in $\mathbf{Z}$ summarizes non-local correlation between the entire primary feature map and location $k$ of the cross modality feature map, as well as the information from the original primary feature map at $k$.

\begin{figure}[t]
	\includegraphics[width=1.0\textwidth]{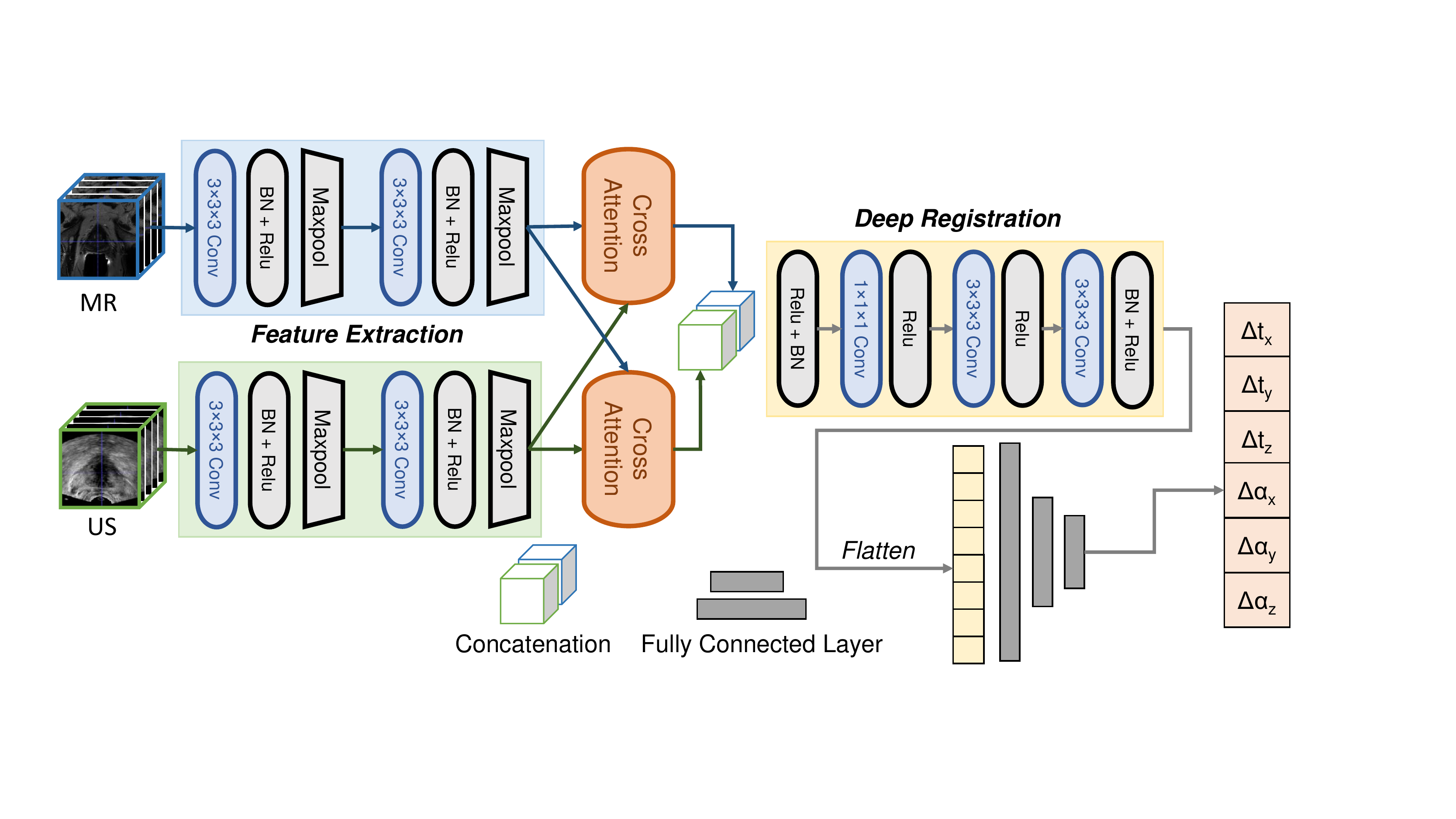}
	\caption{{Overview of the proposed network structure.}}
	\label{fig:network}
\end{figure}

\subsection{Feature Extraction and Deep Registration Modules}

In the proposed network shown in Fig.~\ref{fig:network}, feature extraction modules precede the cross-modal attention block to efficiently represent the input volumes. Each feature extraction module consists of two sets of convolutional and maxpooling layers. 
The deep registration module fuses the concatenated outputs of the cross-modal attention blocks, and predicts the transformation for registration. Other works have used very deep neural networks to automatically learn the complex features of inputs \cite{xie2017aggregated}. However, since the cross-modal attention blocks help determine the spatial correspondence between the two sets of volumes, our registration module can afford to be light weighted. Thus, only three convolutional layers are used to fuse the two feature maps. The final fully connected layers convert the learnt spatial information into an estimated transformation.

\subsection{Implementation Details}

Due to the difficulty in representing the complex image appearances of MRI and TRUS images, surface-based and surface to volume registration methods have been investigated with considerable success \cite{bashkanov2021learning,thomson2020mr,zhang2017recent}. That inspired us to replace the MRI volume with the prostate segmentation label volume in our work. The network remains the same and we only need to set the fixed image input as either MRI volume or segmentation label. The corresponding networks are named as Attention-Reg (image) and Attention-Reg (label), respectively. One advantage of using MRI prostate segmentation is that the binary representation is much more tolerant to image quality and device specificity than MRI volume. Moreover, using segmentation as input can readily extend the proposed method to other imaging modalities, like computed tomography. This implies that while we trained our segmentation guided model on MRI and ultrasound, it may potentially be used on any two modalities.

In this work, we focus on rigid transformation based registration. This decision is determined by the better accessibility of ground truth labels for rigid transformation, and the idea of focusing on network structure comparison only. 
Rigid transformations in this work are performed with 4$\times$4 matrices generated from 6 degrees of freedom \( \theta = \{ \Delta t_x, \Delta t_y, \Delta t_z, \Delta a_x, \Delta a_y, \Delta a_z\}\). These 6 transformation parameters represent translations and rotations along the $x$, $y$, and $z$ directions, respectively. We supervise the network by calculating the MSE (Mean Square Error) between the prediction and the ground truth parameters. 

In our experiments, we included the recent methods of MSReg by Guo \etal~\cite{guo2020deep} and DVNet by Sun \etal~\cite{sun2018towards} as benchmarks.
We used Adam optimizer \cite{kingma2014adam} with maximum of 300 epochs to train all the networks including our proposed Attention-Reg approach. We used a step learning rate scheduler for MSReg training, with initial learning rate $5\times10^{-5}$ which then decays to 0.9 every 5 epochs, as suggested in \cite{guo2020deep}. For DVNet, we used the same scheduler but with initial learning rate adjusted to $1\times10^{-3}$. The models were trained on a NVIDIA DGX-1 deep learning server with batch size of 16 for MSReg, and 8 for our proposed network. The testing phase and runtime benchmark were performed on a work station equipped with NVIDIA GeForce RTX 2080 Ti and AMD Ryzen 9 3900X. Both the proposed and the MSReg methods were implemented in Python using the open source PyTorch library \cite{pytorch}. Our implementation of the proposed Attention-Reg is available at: \url{https://github.com/DIAL-RPI/Attention-Reg}.

\section{Experiments and Results}

\subsection{Dataset and Preprocessing}

In this work, we used 528 cases of MRI-TRUS volume pair for training, 66 cases for validation, and 68 cases for testing. Each case contains a T2-weighted MRI volume and a 3D ultrasound volume. Each MRI volume has 512$\times$512$\times$26 voxels with 0.3mm resolution in all directions. The ultrasound is reconstructed from an electro-magnetic tracked freehand 2D sweep of the prostate. The training set was generated afresh for every training epoch to boost model robustness. On the contrary, the validation set consists of 5 pre-generated initialization matrices for each case, resulting in 330 total samples. The reason for not regenerating new validation sets every epoch is to monitor the epoch-to-epoch performance in a more stable manner. For testing, we generated 40 random initialization matrices for each case. The same test set is used for all experiments.

We measured the image registration performance using surface registration error (SRE). To accurately generate a dataset of with known SRE for training and validation, we perturbed each ground truth transformation parameter randomly within the range of 5mm of translation or 6 degrees of rotation, and then scale the perturbation to a random SRE within the desired range.

\subsection{Experimental Results}

\begin{table}[t]
\caption{\label{table:result0}Performance comparison between Attention-Reg and similarity-based iterative registration methods.}
\centering
\begin{tabular}{l|c|cc}
	\hline
\textbf{Method}  & \textbf{Initialization}    & \textbf{Result SRE (mm)}  \\ \cline{1-3} 
Mutual Information~\cite{maes1997multimodality}  & \multirow{4}{*}{8mm} & 8.96$\pm$1.28  \\
SSD MIND~\cite{HeinrichMIND} &    & 6.42$\pm$2.86         \\
Attention-Reg (img) &             & \bf{3.63$\pm$1.86}               \\
Attention-Reg (label) &             & \bf{3.54$\pm$1.91}                 \\\cline{1-3} 
Mutual Information~\cite{maes1997multimodality}  & \multirow{4}{*}{16mm} & 10.07$\pm$1.40  \\
SSD MIND~\cite{HeinrichMIND} &    & 6.62$\pm$2.96         \\
Attention-Reg (img) &             & \bf{4.17$\pm$2.14}               \\
Attention-Reg (label) &             & \bf{4.06$\pm$2.10}                 \\
\hline
\end{tabular}
\end{table}

We first compared our approach to classical iterative registration methods. Table~\ref{table:result0} summarizes the comparison of our method and traditional iterative registration approaches, including mutual information \cite{maes1997multimodality} and MIND \cite{HeinrichMIND} based registration as in \cite{haskins2020deep}.
We tested our result on two sets of initial registrations. One set is initialized at SRE=8mm, and the other set is initialized at SRE=16mm. The results of our proposed models are averaged from 6,800 test samples, with 68 cases of MRI-TRUS volume pair and 100 initialization matrices each. We used this large test set to improve the robustness of the evaluation. 
Attention-Reg (img) stands for the the registration result of our proposed network with MRI volume as the input of fixed image, whereas Attention-Reg (label) uses MRI prostate segmentation label as the fixed image. In both test scenarios, our methods outperformed the traditional approaches significantly ($p<$0.001 under $t$-test). It is also worth noting that when using MRI prostate segmentation as the fixed image, the performance of our network is slightly improved with statistical significance ($p<$0.001 under $t$-test).

\begin{table}[t]
\caption{\label{table:result1} Performance comparison between Attention-Reg, MSReg \cite{guo2020deep}, and DVNet \cite{sun2018towards}. Both parameter count and runtime were measured per stage. 
SRE values are in mm.}
\centering
\begin{tabular}{l|c|cc|r|c}
	\toprule
\textbf{Method}  & \textbf{Initial.}  & \textbf{Stage 1} & \textbf{Stage 2} & \textbf{\#Parameters} & \textbf{Runtime} \\ \cline{1-6}
DVNet~\cite{sun2018towards}    & \multirow{4}{*}{{[}0,20mm{]}} & 4.77$\pm$3.17                    & -          &   5,275,832         & 3ms        \\
MSReg~\cite{guo2020deep}    &  & 4.75$\pm$2.63                    & 4.04$\pm$2.30          &   16,106,076         & 6ms        \\
Attention-Reg (image)                   &                           & 4.50$\pm$2.58             & 3.71$\pm$1.99          &  1,248,777        & 3ms        \\
Attention-Reg (label) &                          & \bf{4.44$\pm$2.32}           & \bf{3.60$\pm$2.01}          &   1,248,777         &   3ms    \\

\midrule 
\midrule 

Feature-Reg (image)                   &     \multirow{2}{*}{{[}0,20mm{]}}                      & 5.14$\pm$2.58             & -          &  1,244,393        & 3ms        \\
Feature-Reg (label) &                          & 5.22$\pm$2.81           &    -       &   1,244,393         &   3ms    \\
\bottomrule
\end{tabular}
\end{table}

Table~\ref{table:result1} lists the results of our method and other end-to-end rigid registration techniques, including MSReg by Guo \etal~\cite{guo2020deep} and DVNet by Sun \etal~\cite{sun2018towards}. The ResNeXt structure that Guo \etal~adopted is one of the more advanced variations of CNN \cite{xie2017aggregated}, adding more weight to this comparison. The 2D CNN network in DVNet \cite{sun2018towards} treats 3D volumes as patches of 2D images, a lighter approach in handling 3D volume registration. We tested these networks on 2,720 testing samples, which consists of 68 cases with 40 initialization positions for each case.
To better compare our Attention-Reg with MSReg~\cite{guo2020deep}, which used two consecutive networks to boost performance, we also trained our network twice on two differently distributed training sets. The model for the \(1^{st}\) stage was trained and tested on a generated dataset with initial SRE uniformly distributed within the range of \([0,20mm]\), and the range for the \(2^{nd}\) stage was set to be \(SRE\in[0,8mm]\). The trained networks were concatenated together to form a two-stage registration network. 

As shown in the top part of Table~\ref{table:result1}, our cross-modal attention network outperformed MSReg in both registration stages. Furthermore, the better result was achieved with only 1/10 the number of parameters, and half the runtime. The significantly smaller model and simpler calculation demonstrate that the proposed cross-modal attention block can efficiently capture key features of the image registration task. Again, we observed that the performance of our network with segmentation label as input was consistently better, with significantly reduced SRE when compared to MSReg ($p<$0.001) in both stages.

\begin{figure}[t]
 \includegraphics[width=1.0\textwidth]{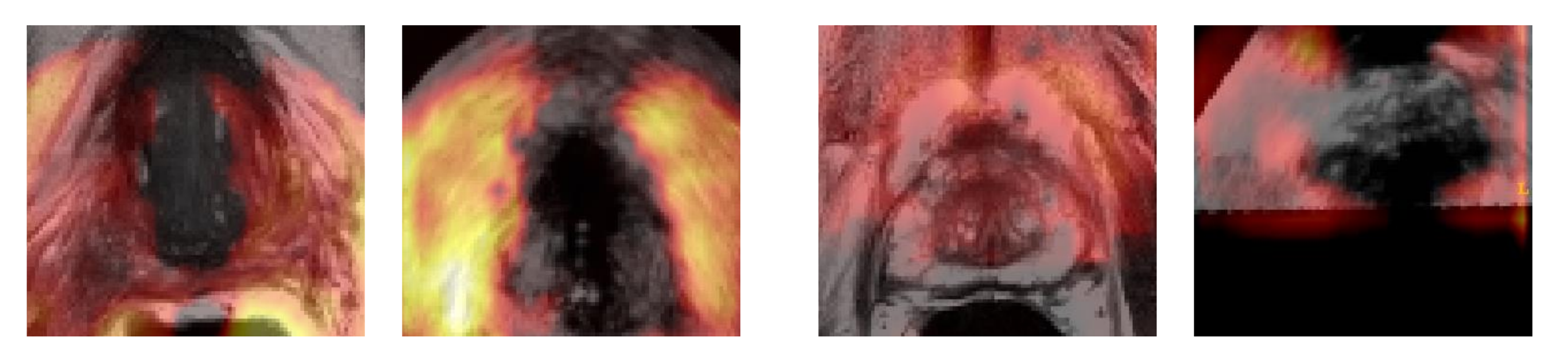}
 \hfill
 \includegraphics[width=1.0\textwidth]{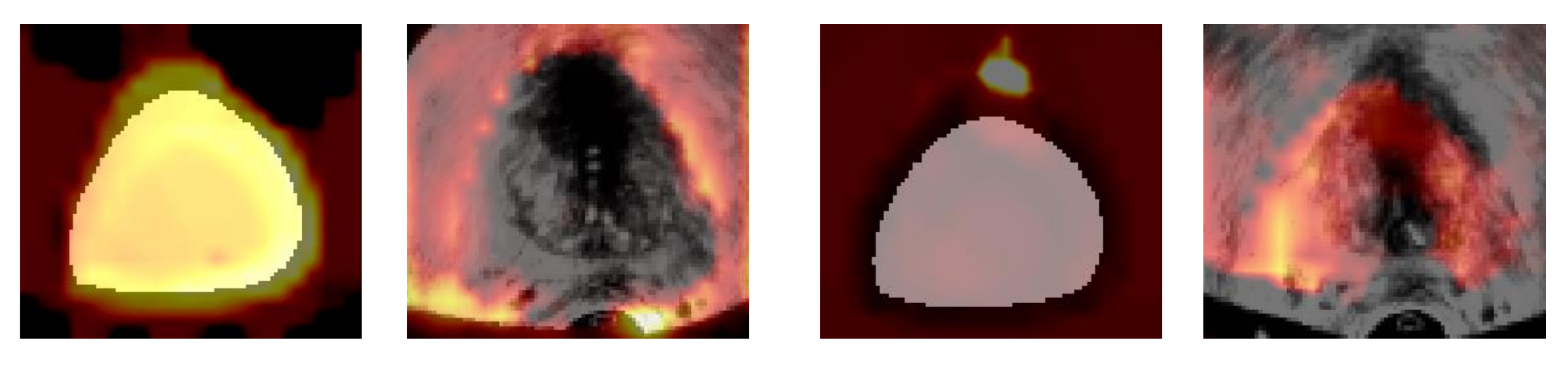}
\caption{Grad-CAM visualization of four pairs of feature maps resulting from the multi-modal attention blocks of \textbf{(top)} Attention-Reg (image) and \textbf{(bottom)} Attention-Reg (label). The image on the left and right in each pair are from the fixed and moving images, respectively.}
\label{fig:visualize_2lines}
\end{figure}

To demonstrate the contribution of the proposed cross-modal attention block, we trained our Attention-Reg network without the attention block, \textit{i.e.}, directly concatenating the outputs of feature extraction modules and feeding to the deep registration module. The results are shown in the bottom half of Table~\ref{table:result1}, which prove the importance of the proposed cross-modal attention block. Without the attention module, the registration performance under both settings was significantly reduced ($p<$0.001 with paired $t$-test). Also, note that without the attention block, using segmentation label as fixed image no longer has an advantage over MRI volume. We speculate that this is also caused by the loss of attention block, which establishes a sensible spatial correlation between the MRI segmentation and the ultrasound volume, as shown in Fig.~\ref{fig:visualize_2lines}.

To help understand the function of cross-modal attention blocks, we employed Grad-CAM~\cite{selvaraju2017grad} to visualize the output of the two multi-modal attention blocks. Similar with Grad-CAM, we used the preceding CNN layer's weight gradient to scale the importance of each feature map channel, and thereby acquired a single volume that represents the output of the multi-modal attention block. 
Fig.~\ref{fig:visualize_2lines} shows the visualization result.
It is apparent that both MRI and ultrasound features are roughly the shape and location of the corresponding ultrasound frame. This means that the network is focusing on the same region of information in both volumes.

\section{Conclusion}
\label{sec:conclusions}
This paper introduced a novel attention mechanism for the task of medical image registration. By comparing the proposed network with other classical methods and purely CNN-based networks up to ten times of its size, we demonstrated the effectiveness of the new cross-modal attention block. To emphasize the importance of prostate boundary, we also quantitatively evaluated the effect of replacing an MRI volume with its segmentation mask as network input. Our proposed methods have led to significant improvements in image registration accuracy over the previous registration methods.
Through feature map visualization, we observed that the network indeed extracted meaningful features to guide image registration. We expect to see our methods tested out in other medical image registration settings in the future with such improvement in accuracy and efficiency, and interpretability.

%
%
\bibliographystyle{splncs04}
\bibliography{references}

\end{document}